%% file: ecai2025.tex
\documentclass{ecai} 

\usepackage{latexsym}
\usepackage{amssymb}
\usepackage{amsmath}
\usepackage{amsthm}
\usepackage{booktabs}
\usepackage{enumitem}
\usepackage{graphicx}
\usepackage{color}

\usepackage{booktabs} 
\usepackage{multirow}
\usepackage{makecell}

\newcommand{\BibTeX}{B\kern-.05em{\sc i\kern-.025em b}\kern-.08em\TeX}

\begin{document}


\begin{frontmatter}



	\title{Seeing the Unseen: Mask-Driven Positional Encoding and Strip-Convolution Context Modeling for Cross-View Object Geo-Localization}


	\author[A]{\fnms{Shuhan}~\snm{Hu}}
	\author[A]{\fnms{Yiru}~\snm{Li}}
	\author[A]{\fnms{Yuanyuan}~\snm{Li}}
	\author[A]{\fnms{Yingying}~\snm{Zhu}\footnote{Corresponding author.}}

	\address[A]{College of Computer Science and Software Engineering, Shenzhen University}


	\input{sec/0_abstract}

\end{frontmatter}


\input{sec/1_intro}


\input{sec/2_related_work}


\input{sec/3_method}

\input{sec/4_experience}

\input{sec/5_conclusion}







\bibliography{ref}

\end{document}

%% file: sec/0_abstract.tex
\begin{abstract}

    Cross-view object geo-localization enables high-precision object localization through cross-view matching, with critical applications in autonomous driving, urban management, and disaster response. However, existing methods rely on keypoint-based positional encoding, which captures only 2D coordinates while neglecting object shape information, resulting in sensitivity to annotation shifts and limited cross-view matching capability. To address these limitations, we propose a mask-based positional encoding scheme that leverages segmentation masks to capture both spatial coordinates and object silhouettes, thereby upgrading the model from "location-aware" to "object-aware." Furthermore, to tackle the challenge of large-span objects (e.g., elongated buildings) in satellite imagery, we design a context enhancement module. This module employs horizontal and vertical strip convolutional kernels to extract long-range contextual features, enhancing feature discrimination among strip-like objects. Integrating MPE and CEM, we present EDGeo, an end-to-end framework for robust cross-view object geo-localization. 
    Extensive experiments on two public datasets (CVOGL and VIGOR-Building) demonstrate that our method achieves state-of-the-art performance, with a 3.39\% improvement in localization accuracy under challenging ground-to-satellite scenarios.
    This work provides a robust positional encoding paradigm and a contextual modeling framework for advancing cross-view geo-localization research.


\end{abstract}

%% file: sec/1_intro.tex
\section{Introduction}
\label{sec:intro}

Cross-view object geo-localization (CVOGL) is a critical task that addresses the challenge of precisely locating specific objects when direct GPS signals are weak or unavailable\cite{friedland2010multimodal}, such as in urban canyons. The core idea of CVOGL is to identify a user-specified object in a reference image (typically a geo-tagged satellite image) based on its indication in a query image (often a street-level or UAV image). By utilizing the relative positional relationship of the query object within the reference image, alongside the geographical metadata of the reference image, CVOGL can determine the precise geographic coordinates of the target object. This capability for high-accuracy object localization makes CVOGL highly valuable across a range of real-world applications, including but not limited to autonomous driving\cite{hane20173d,xia2021cross,kumar2023survey}, robotic navigation\cite{mcmanus2014shady}, urban management\cite{yao2022multimodal,sun2019streaming}, post-disaster rescue operations\cite{li2023spatiotemporal,dahlke2024seamless,garcia2021computer,sogi2024disaster}, and GPS spoofing defense\cite{jiang2024seek+,jiang2023seek}.

\begin{figure}[t]
	\centering
	\includegraphics[width=\linewidth]{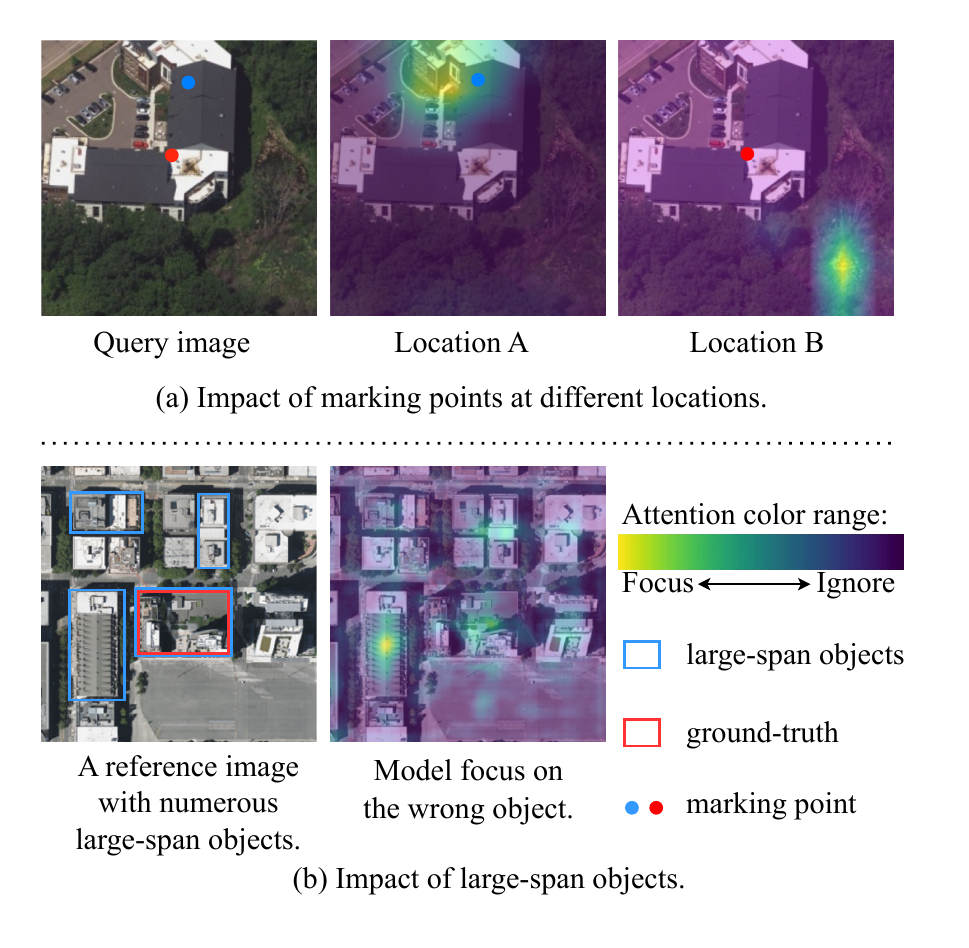}
	\caption{Impact of marker location and large-span objects on the CVOGL task.}
	\label{fig:motivation}
\end{figure}

Current CVOGL detectors typically rely on a single user click to indicate the target in the query image; the click is converted into a keypoint-based positional encoding (KPE) that is concatenated with the image features. Although simple, KPE conveys only 2D coordinates, ignoring the extent, outline, and orientation of the target, and it is notoriously sensitive to slight annotation shifts, leading to a restricted perceptual understanding by the model, as shown in Figure~\ref{fig:motivation}a. First, KPE relies solely on coordinate information of marking points without characterizing the shape of the object, resulting in weak perception of query objects. Second, KPE cannot stably identify the location of objects, making the model performance highly sensitive to coordinate shifts of marking points. Finally, the lack of shape information makes it more challenging to match objects across drastically different viewpoints (street view to satellite), as the appearance around a single point can vary significantly between views. To address these limitations, we recognize the need to obtain both precise position and shape information of objects. Drawing inspiration from image segmentation tasks\cite{ye2024sgbev,kirillov2023segment,minaee2022survey,rahman2024emcad}, which divide images into non-intersecting areas and locate objects at the pixel level, we propose mask-based positional encoding (MPE). This approach leverages segmentation masks to capture refined positional and shape information of query objects, effectively evolving the model from merely "location-aware" to "object-aware".

Furthermore, existing methods adapt general object detection techniques\cite{gui2024rsodreview,fan2024small,li2024lrfpn,liu2024tiny,fu2024sodet} without adequately leveraging the unique characteristics of satellite imagery. Satellite images frequently contain numerous large-span objects (objects with aspect ratios greater than 1.5, such as elongated buildings and roads), as shown in Figure~\ref{fig:motivation}b. On the one hand,  these characteristics remain largely unexplored in current approaches; on the other hand, conventional backbones that rely on small square kernels struggle to capture the long-range horizontal and vertical context required to distinguish one strip-like object from another. 
Thus, large-span objects affect the performance of the model.
Research \cite{yuan2025striprcnnlargestrip} has shown that strip convolution with long kernels can better extract features and improve localization accuracy for such objects. Based on this insight, we propose a Context Enhancement Module (CEM) that employs strip convolution with long kernel design in horizontal and vertical directions to effectively model large-span context characteristics, significantly improving feature discrimination between large-span objects in satellite images.

We propose enhanced detection geo-localization (EDGeo) by integrating the proposed MPE and CEM together. In the EDGeo method, we first use the MPE Generator to generate MPE based on the query image and marking points. Next, we fuse the query image with MPE and send it to a feature extractor to extract the query features. Correspondingly, the reference image is fed into a feature extractor to extract reference features. After the feature fusion module, we fuse the query features with the reference features. Finally, we feed the fused features into CEM for feature enhancement to obtain the final features. The final features will be sent to the detection head for bbox prediction. We validate the effectiveness of the EDGeo through extensive experiments on the CVOGL dataset and VIGOR-Building dataset, achieving state-of-the-art performance with significant improvements over existing methods.

The key contributions of our work are as follows:
\begin{itemize}
    \item We introduce a segmentation-driven mask for cross-view object geo-localization. The mask-based positional encoding (MPE) scheme that embeds both the precise location and the full silhouette of the query object. MPE equips the detector with rich shape cues, greatly reducing sensitivity to click jitter and enabling robust matching across extreme viewpoint changes.

	\item  To exploit the elongated objects pervasive in satellite imagery, we design a Context Enhancement Module (CEM), a dual-branch strip-convolution block that applies horizontal $1 \times k$ and vertical $k \times 1$ kernels. This orientation-aware, long-receptive-field design captures extended context along each axis, boosts discrimination among strip-like objects (e.g., roads, runways, long buildings), and strengthens boundary coherence under cluttered backgrounds, leading to markedly improved geo-localization accuracy.

	\item  We present EDGeo, an end-to-end framework that integrates MPE and CEM modules to address both positional encoding stability and contextual information utilization. Our comprehensive experiments on the CVOGL benchmark demonstrate that EDGeo achieves state-of-the-art performance, with particular improvements of 5.44\% in challenging ground-to-satellite scenarios.
\end{itemize}

%% file: sec/2_related_work.tex
\section{Related Work}
\label{sec:related-work}

\subsection{Cross-view Image Geo-localization}

Cross-view geo-localization \cite{zhu2021geographic,toker2021coming,tian2021uav,guo2022fusing,li2023multi,zhang2024aligning} refers to the task of identifying the image most similar to a given query image within a database of geotagged reference images, thus determining the geographical location of the query image. Cross-view geo-localization technology enables accurate prediction of the geographic location of the query image.

Researchers have made significant contributions to the central task of cross-view geo-localization, which focuses on image retrieval to establish spatial correspondences between images of the same scene captured from different viewpoints or conditions\cite{haigang2022overview}. Hu \cite{hu2018cvm} developed CVM-Net, which incorporates a weighted soft boundary ranking loss function to accelerate training and improve match accuracy. Shi \cite{shi2019spatial} proposed SAFA to address substantial viewpoint differences between ground and aerial images through a two-stage approach: initially aligning the image domains via polar coordinate transformation, followed by a spatial attention mechanism to further minimize content dependency differences, enhancing the accuracy and stability of cross-view geo-localization. Yang \cite{yang2021cross} introduced L2LTR, which uses Transformers' self-attention to capture global dependencies and improve interlayer information flow via cross-attention, achieving notable improvements in accuracy and generalization. Zhu \cite{zhu2022transgeo} presented TransGeo, using the global modeling capacity of Transformers and explicit positional encoding, alongside an attention-guided non-uniform cropping strategy to lower computational costs. Deuser \cite{deuser2023sample4geo} proposed Sample4Geo, employing a contrastive learning framework and symmetric InfoNCE loss function to effectively utilize hard negatives, boosting cross-view geo-localization performance while simplifying training.

However, the CVGL methods are image-level auxiliary geo-localization solutions that cannot achieve more accurate geo-localization for objects in images.

\begin{figure*}[htb]
	\centering
	\includegraphics[width=0.9\linewidth]{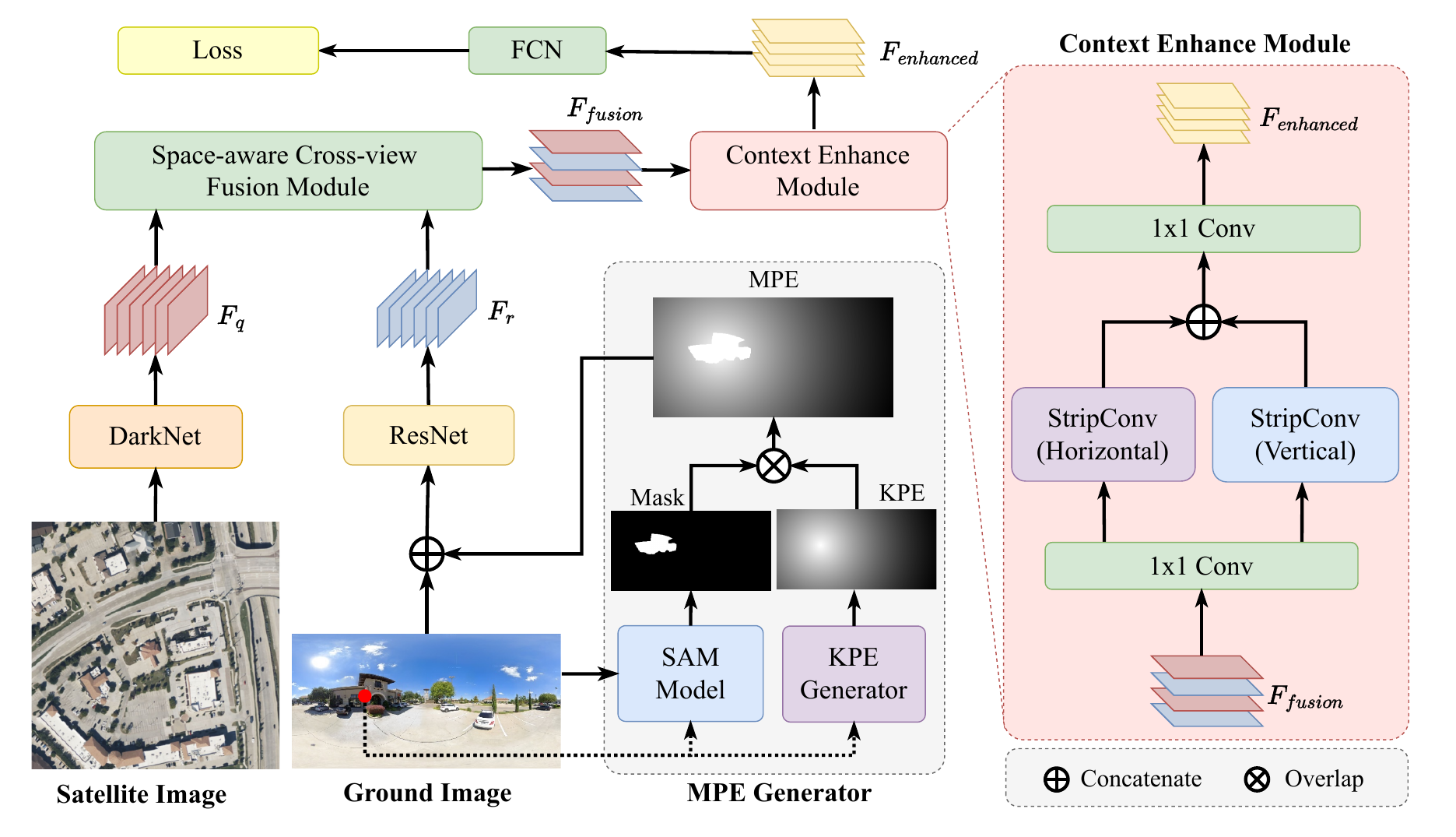}
	\caption{
		Overall architecture of EDGeo model. For a given sample, we first generate the corresponding MPE. Subsequently, the query image is input to the EDGeo model together with the generated MPE. Next, We extract features from the joint query image and MPE. At the same time, the reference image extracts the reference features. Subsequently, the query features and reference features are aggregated through the fusion module and then sent to CEM for feature enhancement to output the final features. Finally, the detection head performs object detection and calculates the loss.
	}
	\label{fig:edgeo-model}
\end{figure*}

\subsection{Cross-view Object Geo-localization}

Cross-view object geo-localization refers to the task of locating a query object in a reference image, where the query object is marked in the query image. In addition, the query image and the reference image are captured from the same location. Sun \cite{sun2023cross} were the first to propose the cross-view object geo-localization task. At the same time, they also proposed DetGeo, which uses spatial attention mechanism to align the features of the query image with those of the reference image, thereby guiding the detection of the query target in the reference image.

%% file: sec/3_method.tex
\section{Method}

\subsection{Problem Formulation}

The cross-view object geo-localization problem can be described as follows: given a dataset $X=\{x_i\}^n_{i=1}=\{q_i, r_i, p_i, b_i\}^n_{i=1}$, consisting of $n$ samples, where each sample includes a query image $q_i$, a reference image $r_i$, and a object marking point $p_i$. The query object is identified in the query image $q_i$ by the query object identifier $p_i$, and represented in the reference image $r_i$ by a bounding box $b_i$. The object identifier $p_i$ is defined by coordinates $(x_{p_{i}},y_{p_{i}})$ in the query image, while the bounding box $b_i$ is represented by its center coordinates, width and height $(x_{b_i}, y_{b_i}, h_{b_i}, w_{b_i})$. The formal definition of this problem is defined as follows:

\begin{eqnarray}\label{eq:task-definition}
	{q_i, r_i, p_i} \mapsto b_i
\end{eqnarray}

\subsection{Overview}

EDGeo consists mainly of four parts: MPE generator, dual-branch image encoder, feature fusion module, and CEM. The model structure is shown in Figure~\ref{fig:edgeo-model}. First, we need to generate a mask map of the query object $m_i$ according to the object marking point $p_i$ in the query image $q_i$. Then we calculate the value of the background part of the mask map $m_i$ using a distance-based method, resulting in the final MPE $pe_m$. Subsequently, the dual-branch image encoder completes the feature extraction of the query image $q_i$ and the reference image $r_i$, and outputs the features of the query image and the reference image $F_i^q$ and $F_i^r$, respectively. It is worth mentioning that the query image $q_i$ will be concatenated with MPE $pe_m$ before performing the query feature extraction. Subsequently, the features $F_i^q$ and $F_i^r$ will be input into the feature fusion module\cite{sun2023cross}. And the feature fusion module will output the fused feature $F_i^{fusion}$. The fused feature $F_i^{fusion}$ is then fed into the CEM to enhance the feature, resulting in the final feature map $F_i$. Finally, the final feature $F_i$ will be passed to the detection head and the prediction object detection result $S_i$ will be output. The predictions are bounding boxes $(\hat{x}_i,\hat{y}_i,\hat{h}_i,\hat{w}_i)$ and confidence scores $\hat{p}_i$ on each pixel of the reference image.

\subsection{Mask-based Positional Encoding}

In the CVOGL task, the query object is determined by the object marking point. Since the CVOGL task needs to achieve object-level geo-localization, the precise query object shape (such as building outline) and location information (such as center point coordinates) jointly determine the model's ability to understand the geometric features of the object, which in turn affects the robustness of the model. Existing methods use KPE to represent the location of the object. However, KPE has two limitations. First, KPE only abstracts the query object as point coordinates and completely ignores the shape information of the object. Therefore, the model's perception of the query object is weak, and the understanding of the model may be affected by occlusion. Second, KPE relies on the absolute coordinates of the marking points, while the marking points in actual scenes are easily disturbed by labeling errors. When the coordinates of the marking point are shifted (e.g. the user clicks on a different location of the target), the attention of the model will be scattered, resulting in limited robustness of the model.

To solve the above problems, we propose a Mask-based Positional Encoding, which introduces the geometric information of the segmentation mask to achieve finer object position and shape modeling. The MPE generation process is shown in Figure~\ref{fig:edgeo-model}. Specifically, MPE first utilizes the zero-sample generalization ability of the image segmentation model to generate candidate masks with query points as hints. For multiple masks that segmentation model may return (such as object local vs. whole), we design an area compromise selection strategy: filter masks with extreme sizes, retain candidate results with medium area, and make a trade-off between segmentation accuracy and noise suppression. Subsequently, in the background part of the mask map, we use a distance-based method to calculate the KPE value to avoid complete loss of the background information in the query image.

The definition of the distance-based encoding calculation method is defined as follows:
\begin{eqnarray}\label{eq:kpe-equation}
	Pos_k(i, j) = (1 - \frac{||z_k(i, j) - p_k||_2}{const})^2
\end{eqnarray}
Here, $Pos_k$ represents the positional matrix for $p_k$, and $Pos(i, j)$ represents the value of the positional encoding at coordinate $(i, j)$, excluding masked regions. $z_k(i, j)$ indicates whether the point at coordinate $(i, j)$ is a landmark, $||\cdot||_2$ represents the Euclidean distance between $z_k(i, j)$ and $p_k$, and $const$ denotes the diagonal length of the query image. Now, the mask-based positional encoding is successfully generated.

By using the mask information of the query object, we obtain a position-stable position code. When the object marking point can correctly identify the object, the area is not easily affected by the coordinate shift of the object marking point. Although the encoding of the background part will be affected by the position shift of the object marking point, because it only represents the area outside the query object, it has little impact on the object position and shape information. Therefore, the use of MPE can effectively improve the robustness of the model.

\subsection{Context Enhancement Module}

The $1 \times 1$ convolution~\cite{lin2013network} has been widely used by researchers for channel expansion and compression, enhancing model nonlinearity, and facilitating cross-channel information exchange. 
As shown in the Figure~\ref{fig:edgeo-model}, the CEM is relatively simple. It consists of two $1 \times 1$ convolutions and two large strip convolutions in different directions. First, the CEM uses a $1 \times 1$ convolution to enhance the expressiveness of features while maintaining consistent input and output feature dimensions. This step is defined as follows:
\begin{eqnarray}\label{eq:step-1}
	{F_i^{fusion}}'=Conv_{1\times 1}(F_i^{fusion})
\end{eqnarray}
where ${F_i^{fusion}}' \in \mathbb{R}^{C_R \times H_R \times W_R}$ is the output of the $1 \times 1$ convolution $Conv_{1\times 1}$.

Next, ${F_i^{fusion}}'$ is subsequently extracted by both horizontal and vertical stripe convolution. By using orthogonal stripe convolution to capture large-span contextual features in the horizontal and vertical directions of the image, CEM can effectively extract features of large-span objects, enhance the features of objects with closed edges, and thereby enhance the discrimination between objects and backgrounds. This step is defined as follows:
\begin{eqnarray}\label{eq:step-2}
	\begin{cases}
		F^v_i=Conv_v({F_i^{fusion}}') \\
		F^h_i=Conv_h({F_i^{fusion}}')
	\end{cases}
\end{eqnarray}
where $Conv_v$ and $Conv_h$ represent vertical and horizontal strip convolutions with long kernel, respectively. The dimensions of $F^h_i$ and $F^v_i$ are consistent with those of ${F_i^{fusion}}'$.

The CEM is capable of capturing features of large-span objects in the reference image along the horizontal and vertical directions. After these strip convolutions, the two resulting features are concatenated along the channel dimension and then input to an $1 \times 1$ convolution to compress the size of the channel dimension. Through channel compression, the dimensions of the output features of the CEM will be consistent with the dimensions of the input features. The step is defined as follows:
\begin{eqnarray}\label{eq:step-3}
	F_i=Conv_{1\times 1}(Cat(F^h_i, F^v_i))
\end{eqnarray}
where $F_i$ is the final output of the CEM, with the same dimensions as ${F_i^{fusion}}$.

The CEM improves the representation capability of the fused features using two $1 \times 1$ convolutions. By fusing horizontal and vertical features, it fully exploits the global information of the reference image. Additionally, the CEM mitigates the loss of local information from the reference image caused by the spatial attention mechanism. Once the fused features are enhanced, the features $S_i$ are used for subsequent detection of the bounding box.

\subsection{Objective Function}

The loss function\cite{sun2023cross} consists of two components: $L_{geo}$ representing the loss in geo-localization of the object, and $L_{cls}$ representing the loss of classification. Together, they define the complete loss function, which is defined as follows:
\begin{eqnarray}\label{eq:loss-function-total}
	L = L_{geo}+L_{cls}
\end{eqnarray}
The individual definitions of $L_{geo}$ and $L_{cls}$ are provided as follows:
\begin{eqnarray}\label{eq:loss-function-geo}
	\begin{aligned}
		L_{geo}
		 & =\sum_{k=1}^n ((\sigma(x_k) - (x_k^* - \lfloor x_k^* \rfloor))^2 \\
		 & +(\sigma(y_k) - (y_k^* - \lfloor y_k^* \rfloor))^2               \\
		 & +(\log \frac{w_k}{w_a} - \log \frac{w_k^*}{w_a})^2               \\
		 & +(\log \frac{h_k}{h_a} - \log \frac{h_k^*}{h_a})^2)              \\
	\end{aligned}
\end{eqnarray}
In this formula, $\sigma(\cdot)$ represents the sigmoid function. The terms $x_i,y_i,w_i,h_i$ are the predicted values of $x,y,w,h$ for the $i$-th bounding box, while $x^*_i,y^*_i,w^*_i,h^*_i$ are the ground truth values. $w_a$ and $h_a$ denote the width and height of the anchor box. During inference, the predicted $w$ and $h$ representing offset values are converted to absolute pixel coordinates using $w=w_i+w_a$ and $h=h_i+h_a$.
\begin{eqnarray}\label{eq:loss-function-cls}
	L_{cls} = \sum_{i=1}^n o^*_i\log(o_i) + (1-o^*_i)\log(1-o_i)
\end{eqnarray}
Here, $o_i$ is the predicted confidence score for the object at a given position, and $o^*_i$ is the corresponding ground truth label. The value of $o^*_i$ is set to 1 only for the bounding box with the highest IoU with the ground truth bounding box, while all other values are set to 0.

%% file: sec/4_experience.tex
\section{Experiment}

\subsection{Datasets and Metrics}

The dataset used in our experiments is the CVOGL dataset\cite{sun2023cross} and the VIGOR-Building dataset\cite{li2025vigor_dataset}. The CVOGL dataset consists of 5,836 high-resolution satellite images which contain 12,478 object instances, 5,279 street view images, and 5,279 drone images. Additionally, the CVOGL dataset includes two subtasks: cross-view object geo-localization from street-view images to satellite images (denoted as "G $\rightarrow$ S") subtask and from drone aerial images to satellite images (denoted as "D $\rightarrow$ S") subtask. These two subtasks are structurally similar, but the perspectives of the query images are different. Compared to the D $\rightarrow$ S subtask, the G $\rightarrow$ S subtask is more challenging due to the greater perspective differences between street view images and satellite images.

The VIGOR-Building dataset advances cross-view object geo-localization research by extending the VIGOR-GEN framework to address the limitations of conventional datasets, specifically targeting many-to-many object mapping scenarios in real-world urban environments. The VIGOR-Building dataset covers three geographically diverse US cities: Chicago, New York and San Francisco. The dataset uses stratified sampling to ensure representative spatial and architectural variation in ground-level and satellite imagery. It has randomly selected images from these cities to ensure diversity and comprehensive coverage. To facilitate object localization, the dataset annotated the ground images using YOLOv9 and the satellite images using OpenStreetMap. In addition, manual annotations were made to refine the dataset.

In object detection, the intersection over union (IoU) is widely used as an evaluation metric and reflects the overlap ratio between two bounding boxes. In this paper, IoU is also applies to measure the accuracy of various methods in the experiments. Accuracy is the main evaluation metric in this study. The formulas for computing IoU-based Accuracy are shown in Equation~(\ref{eq:accuracy-1}),~(\ref{eq:accuracy-2}) and~(\ref{eq:accuracy-3}):
\begin{eqnarray}\label{eq:accuracy-1}
	acc@k=\frac{1}{n}\sum_{i=1}^{n}\Phi_{i}(k)
\end{eqnarray}
where
\begin{eqnarray}\label{eq:accuracy-2}
	\Phi_{i}(k)=
	\begin{cases}
		1, & if\ IoU(b_{i},b_{i}^{*})>k \\
		0, & else
	\end{cases}
\end{eqnarray}
\begin{eqnarray}\label{eq:accuracy-3}
	IoU(b_{i},b_{i}^{*})=\frac{|b_{i}\cap b_{i}^{*}|}{|b_{i}\cup b_{i}^{*}|}
\end{eqnarray}
In these equations, $b_i$ represents the predicted bounding box, and $b_i^*$ is the ground truth bounding box. $|b_i\cap b_i^*|$ denotes the intersection area and $|b_i\cup b_i^*|$ represents the union area of the two bounding boxes. $k$ is the threshold ratio to determine whether a bounding box is correct. In this study, $acc@0.5$ and $acc@0.25$ are selected as the primary metrics to evaluate all methods.

\subsection{Implementation Details}
\label{sec:implementation_details}


The proposed method is implemented using the PyTorch framework. ResNet-18 and DarkNet-53 networks are used with pre-trained weights on ImageNet-1k. The feature fusion module based on the spatial attention mechanism adopts the QACVFM module proposed by DetGeo. The predefined anchor boxes clustered from the CVOGL dataset (defined in $(w,h)$ format) are: (37, 41), (78, 84), (96, 215), (129, 129), (194, 82), (198, 179), (246, 280), (395, 342), (550, 573). The predefined anchor boxes that clustered from the VIGOR-Building dataset are: (137,82), (144,164), (479,243), (255,537), (73,202), (242,117), (175,359), (259,260), (74,108). The SAM model is used in the MPE generator to obtain the mask map of the object. In CEM, we used stripe convolution with kernel sizes of $1 \times 11$ and $11 \times 1$, respectively. During training, we use the RMSProp optimizer and set the initial learning rate to 0.0001, batch size to 12, and train up to 25 epoches.

In order to transform the existing CVGL method into a method that can be used for CVOGL tasks, we refer to the approach of~\cite{sun2023cross}: by dividing the reference image into multiple small blocks and then matching the image on each small block. After obtaining the candidate matches, we calculate the IoU between the bounding boxes of all candidate patches and the ground-truth bounding box. Finally, the bounding box with the highest IoU among the candidate matches is selected as the predicted bounding box.

\subsection{Performance Comparison with State-of-the-art Methods}

We conduct performance comparison experiments on the CVOGL dataset and VIGOR Building dataset to compare the performance of EDGeo with existing methods, which are shown in Table~\ref{tab:cvogl-main-experience-table} and~\ref{tab:vigor-builiding-main-experience-table}. Considering that there are fewer existing methods for CVOGL tasks, we also compare EDGeo with existing CVGL methods. Although CVGL methods can only target the image patch level, some advanced methods can still achieve good results, such as ConGeo. From the experimental results, we can observe that EDGeo achieved state-of-the-art performance in both the CVOGL dataset and the VIGOR-Building dataset. At the same time, we can also observe that the experimental results on the VIGOR-Building dataset are lower than those on the CVOGL dataset, which shows that the VIGOR-Building dataset is more challenging. At the same time, on the VIGOR-Building dataset, our method can still achieve good results on acc@0.25 indicators. In contrast, the performance of the DetGeo method, which is also based on object detection, is significantly reduced, which also shows that our CEM can effectively utilize the features of satellite images to achieve improved model performance.

\begin{table}[ht]
	\caption{
		Performance comparison with existing methods on the CVOGL dataset. \textbf{Bold} and \underline{underlined}
		values represent the best and second-best performance in each category.
	}
	\centering
	\begin{tabular}{llrrrr}
		\toprule
		\multirow{3.5}*{Task}
		 & \multirow{3.5}*{Method}
		 & \multicolumn{2}{c}{Test}
		 & \multicolumn{2}{c}{Validation}                                                                    \\

		\cmidrule{3-6}
		 &
		 & \makecell{acc@                                                                                    \\0.25}
		 & \makecell{acc@                                                                                    \\0.5}
		 & \makecell{acc@                                                                                    \\0.25}
		 & \makecell{acc@                                                                                    \\0.5} \\

		\midrule

		\multirow{10}*{$D \rightarrow S$}
		 & CVM-Net\cite{hu2018cvm}
		 & 20.14                                 & 3.29              & 20.04             & 3.47              \\

		 & L2LTR\cite{yang2021cross}
		 & 38.95                                 & 6.27              & 38.68             & 3.03              \\

		 & RK-Net\cite{lin2022joint}
		 & 19.22                                 & 2.67              & 19.94             & 3.03              \\

		 & Polar-SAFA\cite{shi2019spatial}
		 & 37.41                                 & 6.58              & 36.19             & 6.39              \\

		 & TransGeo\cite{zhu2022transgeo}
		 & 35.05                                 & 6.37              & 34.78             & 5.42              \\

		 & SAFA\cite{shi2019spatial}
		 & 37.41                                 & 6.58              & 36.19             & 6.39              \\

		 & Sample4Geo\cite{deuser2023sample4geo}
		 & 5.75                                  & 1.21              & 6.18              & 0.56              \\

		 & ConGeo\cite{mi2024congeo}
		 & 34.94                                 & 6.66              & 30.60             & 5.60              \\

		 & DetGeo\cite{sun2023cross}
		 & 61.97                     & 57.66 & 59.81 & 55.15 \\

		 & VAGeo\cite{li2025vageo}
		 & \underline{66.19}                     & \underline{61.87} & \underline{64.25} & \underline{59.59} \\

		 & EDGeo(Ours)
		 & \textbf{69.58}                        & \textbf{63.41}    & \textbf{65.76}    & \textbf{60.02}    \\

		\midrule

		\multirow{10}*{$G \rightarrow S$}
		 & CVM-Net\cite{hu2018cvm}
		 & 4.73                                  & 0.51              & 5.09              & 0.87              \\

		 & L2LTR\cite{yang2021cross}
		 & 10.69                                 & 2.16              & 12.24             & 1.84              \\

		 & RK-Net\cite{lin2022joint}
		 & 7.40                                  & 0.82              & 8.67              & 0.98              \\

		 & Polar-SAFA\cite{shi2019spatial}
		 & 20.66                                 & 3.19              & 19.18             & 2.71              \\

		 & TransGeo\cite{zhu2022transgeo}
		 & 21.17                                 & 2.88              & 21.67             & 3.25              \\

		 & SAFA\cite{shi2019spatial}
		 & 22.20                                 & 3.08              & 20.59             & 3.25              \\

		 & Sample4Geo\cite{deuser2023sample4geo}
		 & 6.75                                  & 1.61              & 7.04              & 1.08              \\

		 & ConGeo\cite{mi2024congeo}
		 & 34.94                                 & 6.66              & 30.60             & 5.60              \\

		 & DetGeo\cite{sun2023cross}
		 & 45.43                     & 42.24 & 46.70 & 43.99 \\

		 & VAGeo\cite{li2025vageo}
		 & \underline{48.21}                     & \underline{45.22} & \underline{47.56} & \underline{44.42} \\

		 & EDGeo(Ours)
		 & \textbf{50.87}                        & \textbf{46.76}    & \textbf{49.3}     & \textbf{45.72}    \\

		\bottomrule
	\end{tabular}
	\label{tab:cvogl-main-experience-table}
\end{table}

\begin{table}[ht]
	\centering
	\caption{
		Performance comparison with existing methods on the VIGOR-Building dataset.
	}
	\begin{tabular}{lrrrr}
		\toprule

		\multirow{3.5}*{Method}
		 & \multicolumn{2}{c}{Test}
		 & \multicolumn{2}{c}{Validation}                                                           \\

		\cmidrule{2-5}

		 & \makecell{acc@                                                                           \\0.25}
		 & \makecell{acc@                                                                           \\0.5}
		 & \makecell{acc@                                                                           \\0.25}
		 & \makecell{acc@                                                                           \\0.5} \\

		\midrule

		L2LTR\cite{yang2021cross}
		 & 12.93                          & 1.52             & 13.01             & 1.73             \\

		RK-Net\cite{lin2022joint}
		 & 5.78                           & 0.78             & 8.33              & 0.78             \\

		TransGeo\cite{zhu2022transgeo}
		 & 7.27                           & 1.47             & 5.51              & 0.91             \\

		Sample4Geo\cite{deuser2023sample4geo}
		 & 4.96                           & 0.74             & 6.99              & 0.74             \\

		ConGeo\cite{mi2024congeo}
		 & 20.03              & 2.69             & 22.12 & 3.08             \\

		DetGeo\cite{sun2023cross}
		 & \underline{53.95}                          & \underline{34.68} & \underline{53.7}             & \underline{34.82} \\

		VAGeo\cite{li2025vageo}
		 & 37.74                          & 26.07 & 40.46             & 29.09 \\

		EDGeo(Ours)
		 & \textbf{80.35}                 & \textbf{49.33}   & \textbf{78.79}    & \textbf{53.11}   \\

		\bottomrule
	\end{tabular}
	\label{tab:vigor-builiding-main-experience-table}
\end{table}

\subsection{Ablation Study}

To examine the effects of various modules and parameters in our method, we conduct ablation experiments.

\subsubsection{Ablation Study for Core Components}

\begin{table}[ht]
	\centering
	\caption{
		Ablation study on CEM and MPE.
	}
	\begin{tabular}{lccrrrr}
		\toprule

		\multirow{3.5}*{\makecell{Dataset/                                                            \\Task}}
		 & \multirow{3.5}*{MPE}
		 & \multirow{3.5}*{CEM}
		 & \multicolumn{2}{c}{Test}
		 & \multicolumn{2}{c}{Validation}                                                             \\

		\cmidrule{4-7}

		 &
		 &
		 & \makecell{acc@                                                                             \\0.25}
		 & \makecell{acc@                                                                             \\0.5}
		 & \makecell{acc@                                                                             \\0.25}
		 & \makecell{acc@                                                                             \\0.5} \\

		\midrule

		\multirow{4}*{\makecell{VIGOR-                                                                \\Building}}
		 &
		 &
		 & 13.99                          & 4.2               & 12.42             & 5.23              \\

		 & \checkmark
		 &
		 & 20.28                          & 7.69              & 12.42             & 4.58              \\

		 &
		 & \checkmark
		 & \underline{39.16}              & \underline{13.29} & \underline{32.68} & \underline{13.07} \\

		 & \checkmark
		 & \checkmark
		 & \textbf{46.15}                 & \textbf{18.88}    & \textbf{43.79}    & \textbf{18.95}    \\

		\bottomrule
	\end{tabular}
	\label{tab:ablation-module}
\end{table}

To verify the validity of our model, we conduct ablation experiments on the CVOGL dataset and the VIGOR-Building dataset. The detailed results are shown in Table~\ref{tab:ablation-module}. When the CEM was removed from the model, the performance decreased significantly. This shows that the CEM can effectively improve the discrimination of the object from the background and improve the model performance by extracting the features of the large-span object as well as the directional features. The effect of MPE on the robustness of the model will be demonstrated in subsequent experiments. Although the contribution of MPE to the performance of the model is not obvious, it can produce a synergy effect when combined with CEM: after the model captures the query target through MPE, it can better distinguish the query object from other objects through CEM, and suppress the false detection of similar objects, thereby improving the performance of the model. The impact of MPE on the robustness of the model will be demonstrated in subsequent experiments.

\subsubsection{Ablation Study for Strip Convolution Kernel}

We perform some experiments on the size of the convolutional kernels for stripe convolutions, adjusting the length of the convolutional kernels from 7 to 19. We show the effect of different convolutional kernel sizes on the model performance using metrics acc@0.25 and acc@0.5 in the CVOGL dataset. The experimental results are shown in Table~\ref{tab:kernel-size-analysis}. From the experimental results, we can observe that when the kernel size is 11, it performs stably under both subtasks and achieves optimal or suboptimal performance under all metrics. From the experimental performance, the convolutional kernel is too long or too short to achieve the best performance of the model. We believe that long convolutional kernels lead to receptive field redundancy, which is sensitive to background information noise; too short convolutional kernels cannot capture long-distance features, resulting in local detail loss. Moderate convolution kernels can make an effective trade-off between receptive fields and noise suppression.

We perform some experiments on the size of the convolutional kernels for stripe convolutions, adjusting the length of the convolutional kernels from 7 to 19. We show the effect of different convolutional kernel sizes on the model performance using the metrics acc@0.25 and acc@0.5 in the CVOGL dataset. The experimental results are shown in Table~\ref{tab:kernel-size-analysis}. From the experimental results, we can observe that when the kernel size is 11, it performs stably under both subtasks and achieves optimal or suboptimal performance under all metrics. From the experimental performance, the convolutional kernel is too long or too short to achieve the best performance of the model. We believe that long convolutional kernels lead to receptive field redundancy, which is sensitive to background information noise; too short convolutional kernels cannot capture long-distance features, resulting in local detail loss. Moderate convolution kernels can make an effective trade-off between receptive fields and noise suppression.

\begin{table}[ht]
	\centering
	\caption{
		Analysis for kernel size of strip convolution in CEM on the CVOGL Dataset.
	}
	\begin{tabular}{ccrrrr}
		\toprule

		\multirow{3}*{Task}
		 & \multirow{3}*{\makecell{Kernel                                                                                 \\size}}
		 & \multicolumn{2}{c}{Test}
		 & \multicolumn{2}{c}{Validation}                                                                                 \\

		\cmidrule{3-6}

		 &
		 & \makecell{acc@                                                                                                 \\0.25}
		 & \makecell{acc@                                                                                                 \\0.5}
		 & \makecell{acc@                                                                                                 \\0.25}
		 & \makecell{acc@                                                                                                 \\0.5} \\

		\midrule

		\multirow{6}*{$G \rightarrow S$}
		 & 17                             & \underline{48.92} & \underline{45.22} & 46.91             & 43.45             \\
		 & 15                             & 47.69             & 43.58             & 44.96             & 41.17             \\
		 & 13                             & 48.10             & 43.27             & 48.00             & 42.90             \\
		 & 11                             & \textbf{50.87}    & \textbf{46.76}    & \underline{49.30} & \textbf{45.72}    \\
		 & 9                              & 46.04             & 42.96             & 46.48             & 42.15             \\
		 & 7                              & 48.10             & 44.40             & \textbf{49.40}    & \underline{45.50} \\

		\midrule

		\multirow{6}*{$D \rightarrow S$}
		 & 17                             & 62.08             & 54.14             & 58.72             & 52.87             \\
		 & 15                             & \underline{68.76} & \underline{61.56} & \textbf{66.85}    & \textbf{60.13}    \\
		 & 13                             & 67.21             & 59.61             & 64.36             & 57.75             \\
		 & 11                             & \textbf{69.58}    & \textbf{63.41}    & \underline{65.76} & \underline{60.02} \\
		 & 9                              & 65.06             & 59.61             & 65.01             & 58.18             \\
		 & 7                              & 65.36             & 58.79             & 63.06             & 57.75             \\

		\bottomrule
	\end{tabular}
	\label{tab:kernel-size-analysis}
\end{table}

\subsection{Robustness Analysis}

\begin{figure}[ht]
	\centering
	\includegraphics[width=0.7\linewidth]{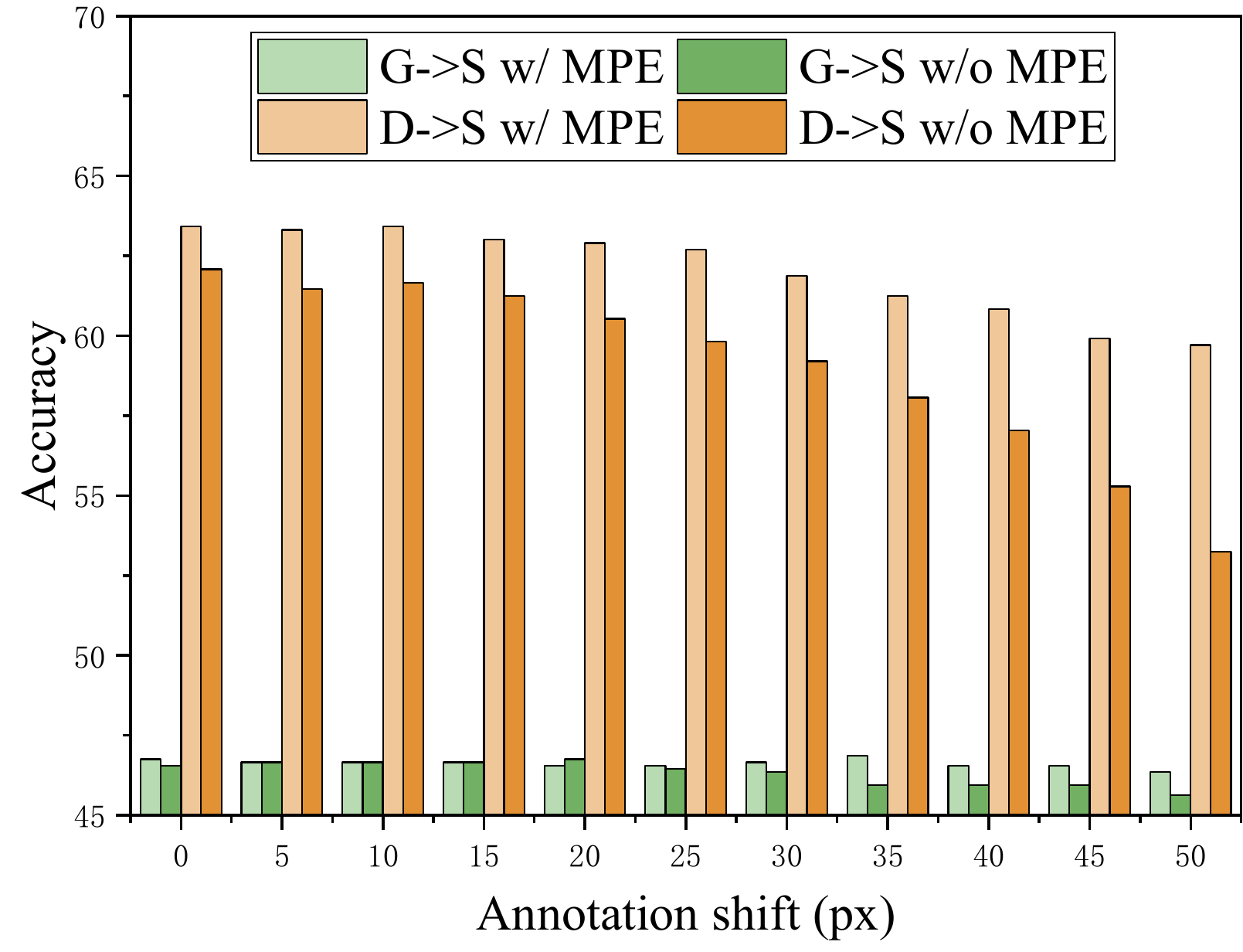}
	\caption{Performance comparison without / with MPE.}
	\label{fig:robustness-experience}
\end{figure}

\begin{figure*}[htb]
	\centering
	\includegraphics[width=0.9\linewidth]{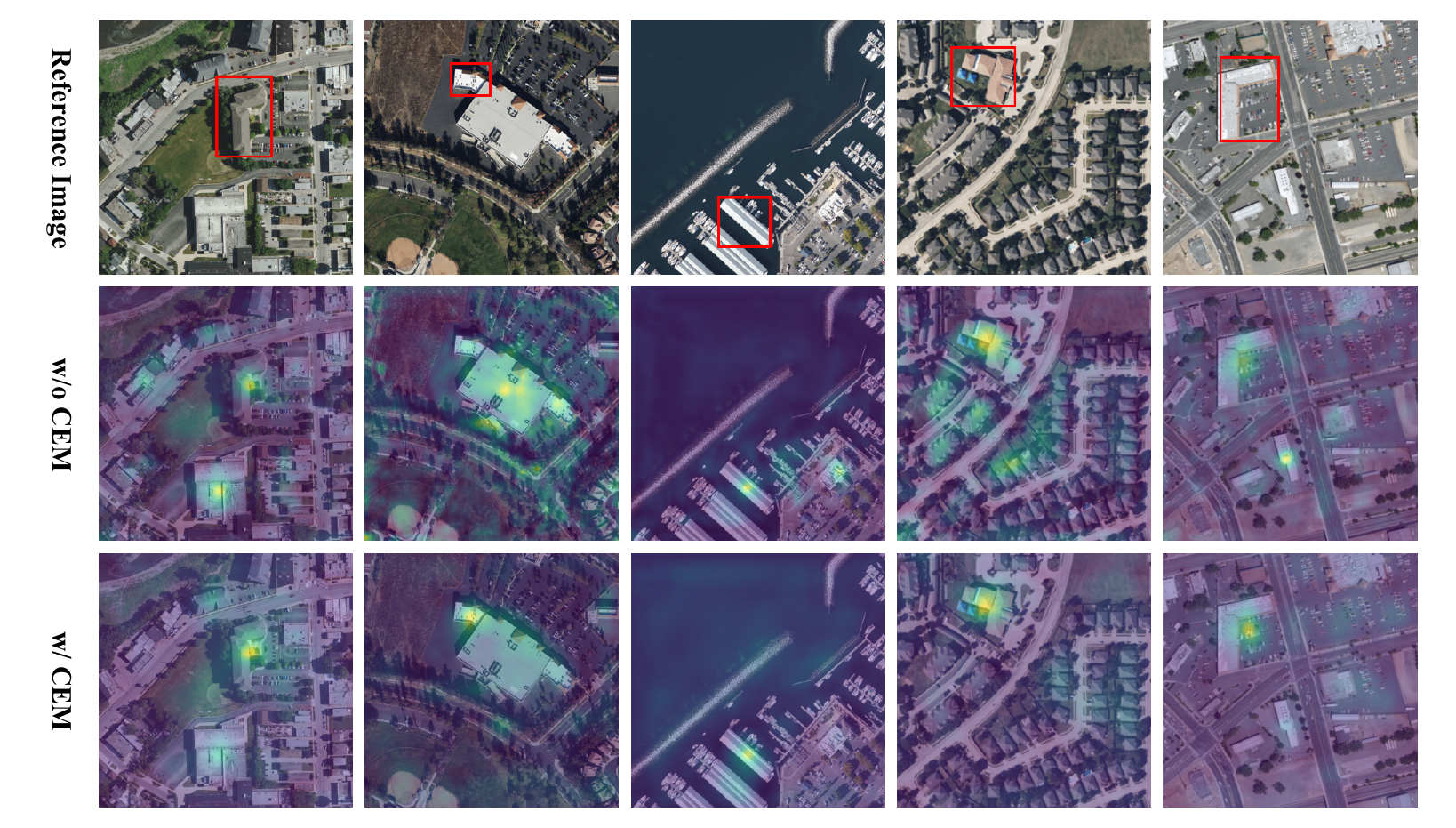}
	\caption{Visualization of model for CEM. We compare the changes in the model's attention to the reference image with/without using CEM. The red bounding box shows where the query object is in the reference image. "w/o CEM" and "w/ CEM" represents our visualization of the model without/with the CEM, respectively.}
	\label{fig:cem-visualization}
\end{figure*}

\begin{figure*}[htb]
	\centering
	\includegraphics[width=0.8\linewidth]{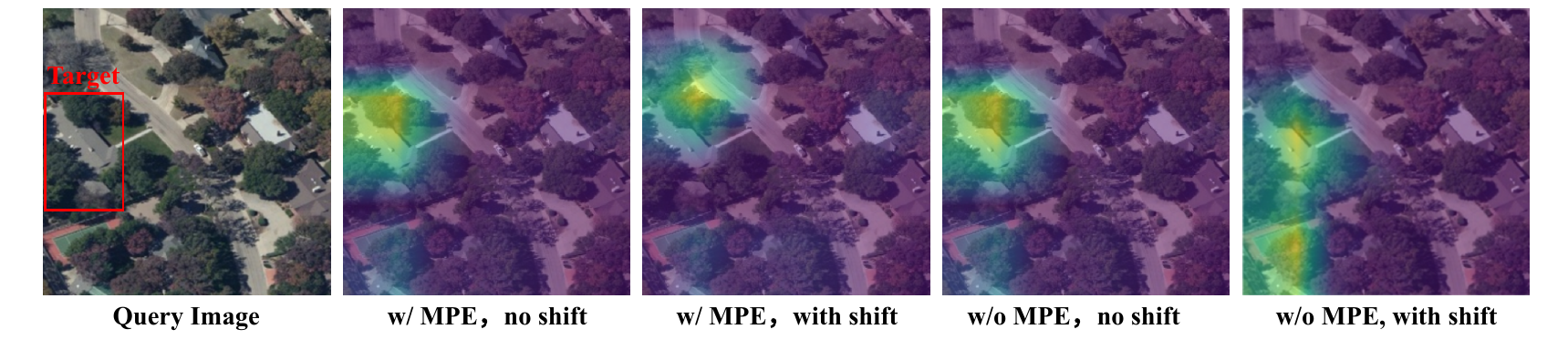}
	\caption{Visualization of model for MPE. We compare the changes in the model's attention to the query image with/without the location shift of the marker points. "w/o MPE" and "w/ MPE" represents our visualization of the model without/with the MPE, respectively. "no shift" represents using the original marking point from the dataset, while "with shift" represents using the marking point that is shifted from the original point.}
	\label{fig:mpe-visualization}
\end{figure*}

In order to verify the effectiveness of MPE in improving the robustness of the model, we conduct experiments on the CVOGL dataset. We add pixel-level location shifts to the marking point coordinates and EDGeo, and we ensure that the shifted points still fell on the object through the mask map. We observe the performance changes of the model before and after using KPE and MPE under different degrees of shift of the marking point coordinates. The specific experimental results are shown in Figure~\ref{fig:robustness-experience}. From the figure, we can see that with the increase in the marking point coordinate offset, the performance of the model on the $D \rightarrow S$ subtask and the $G \rightarrow S$ subtask gradually decreased, and the larger the marking point coordinate offset, the more severe the performance decline. At the same time, we can also find that using MPE, even if the marking points are shifted, the model performance is more stable compared to not using MPE, indicating that MPE can better improve the robustness of the model. In addition, we can also observe that with an increase in coordinate shift of the marker point, the degree of recovery of model performance is higher when using MPE, indicating that MPE can better suppress the impact of coordinate shift of the marker point.

\subsection{Visualization}

We conduct visualization experiments on the reference image for the proposed CEM and on the query image for MPE. By visualizing the model's attention to the reference image and query image, we can clearly see the effects of MPE and CEM.

From the CEM visualization results, which is shown in Figure~\ref{fig:cem-visualization}, we can see that after using the CEM module, the model can better identify the query object and distinguish it from other nearby objects and roads; at the same time, the model can effectively extract features of large-span objects to capture the query object. For example, in the third column, "w/CEM" image can better distinguish the building on the right side of the graph compared to "w/o CEM" image, improve discrimination, and avoid wrong detections.

From the visualization results of MPE, which is shown in Figure~\ref{fig:mpe-visualization}, we can see that when the marking points of the query target are shifted in location, the model's attention to the query target remains relatively stable when using MPE (comparing "w/MPE, no shift" and "w/MPE, with shift"), although there is also a slight shift; When MPE is not used (comparing "w/o MPE, no shift" and "w/o MPE, with shift"), the model's attention to the query target is significantly dispersed. This shows that MPE can effectively improve the robustness of the model.

%% file: sec/5_conclusion.tex
\section{Conclusion}

In this paper, we propose EDGeo, a novel method for the cross-view object geo-localization task. We propose a novel mask-based positional encoding to increase the robustness of the model. Using mask-based positional encoding, the query object in the query image can be more effectively and robustly identified compared to keypoint positional encoding. Furthermore, to make full use of the reference image information, we introduce a context enhancement module to enhance the aggregated features. This module adds more layout information from the reference image by leveraging its global information, which helps improve the detection of the query object. Our extensive experiments show that our method achieves state-of-the-art performance and demonstrates strong robustness to variations in object marking points.
There is a limitation in this study. The existing image segmentation models still have some incorrect segmentation, which will effect the robustness of model. To address the limitation, we will conduct further research in future work.